\documentclass[]{article}
\usepackage{proceed2e}
\usepackage{algorithm}
\usepackage{algpseudocode}

\usepackage{color}

\usepackage{amsfonts}
\usepackage{amssymb}
\usepackage{amsmath}
\usepackage{graphicx}
\usepackage{booktabs}

\usepackage[compact]{titlesec}
\titlespacing{\section}{3pt}{3pt}{3pt}

\usepackage[round,authoryear]{natbib}
\usepackage{url}

\newcommand{\ud}{\mathrm{d}}


\title{Efficient Bayesian Nonparametric Modelling of Structured \\Point Processes}

\author{ {\bf Tom Gunter}\thanks{\;\; Corresponding authors, in alphabetical order.} \\ 
\And 
{{\bf Chris Lloyd}\textnormal{\textsuperscript{*}}} \\            
\And
{\bf Michael A. Osborne} \\ 
\And 
{\bf Stephen J. Roberts} \\
\AND
\textnormal{Machine Learning Research Group} \\
\textnormal{Department of Engineering Science}\\
\textnormal{University of Oxford} \\
\url{{tgunter, clloyd, mosb, sjrob}@robots.ox.ac.uk}
} 

\newcommand{\xt}{\tilde{x}}
\newcommand{\xtm}{\tilde{x}_m}

\begin{document}

\maketitle

\begin{abstract}
This paper presents a Bayesian generative model for dependent Cox point processes, alongside an efficient inference scheme which scales as if the point processes were modelled independently. We can handle missing data naturally, infer latent structure, and cope with large numbers of observed processes. A further novel contribution enables the model to  work effectively in higher dimensional spaces. Using this method, we achieve vastly improved predictive performance on both 2D and 1D real data, validating our structured approach.
\end{abstract}

\section{\MakeUppercase{Introduction}}\label{sec.introduction}

Point processes are effectively used to model a variety of event data, and have also shown a recent popularity within the Machine Learning community as priors over sets. The most fundamental example of such a stochastic model for random sets is the homogenous Poisson process. This is defined via an intensity which describes the expected number of points found in any bounded region of some arbitrary domain. An inhomogenous Poisson process allows the intensity to vary throughout the domain over which the process is defined. As we do not know the functional form of this intensity given only event data,  another stochastic process is typically used to model it nonparametrically. This is then termed a doubly-stochastic Poisson process, a type of renewal process also known as a Cox process. In our particular construction, we use transformed Gaussian processes to model the intensity functions of the individual dependent point processes, in such a manner as to enable fully nonparametric Bayesian inference \citep{Adams_09_a,Murray_10_a}. While we only explicitly consider the doubly stochastic Poisson process, any general renewal process \citep{Rao_11_a} could be incorporated into the framework we define.

There are many occasions when we have multiple point processes which we expect to be dependent: If the domain is temporal, then an example would be individual clients making trades with a specific financial services provider, or individual customers purchasing items from a specific vendor. If the domain is spatial, we might consider different categories of crime defined over some geographic region.  Defining a flexible model for inter-process dependency structure, alongside an efficient inference scheme allows us to learn the underlying intensity functions which drive the typical behaviour. These can then be used to make more accurate predictions, especially during periods of unobservability for an individual process.

In order to maximise the flexibility of our approach, we specifically assume that the individual intensity functions arise via a weighted summation of convolutions of latent functions with a kernel. Intuitively this means that we take a small number of latent functions, individually smooth and scale them, and then add them together to yield an intensity function. This approach allows a  wide range of intensities to arise from only a few latent functions.

We will present and validate the following novel contributions:
\begin{itemize}
  \setlength{\itemsep}{0pt}
\item The first generative model for dependent Cox process data (Section \ref{sec.model}).
\item An efficient, parallelised inference scheme, which scales  benignly with the number of observed point processes (Section \ref{sec.inference}).
\item A new adaptation of thinning \citep{Lewis_79_a}, which we term `adaptive thinning'. This introduces multiple uniformisation levels over the space, making the model viable for higher dimensional spaces and larger datasets (Section \ref{sec.adaptive}).
\end{itemize}

\section{\MakeUppercase{The Model}}\label{sec.model}

We first formally review the Cox process, before describing the innovative nonparametric Bayesian model outlined in \cite{Adams_09_a} as the Sigmoidal Gaussian Cox Process (SGCP), which allows a full Gaussian process to be used as a prior over an individual intensity function. We then move on to review the convolution process \citep{Alvarez_11_a}, a method of modelling dependent functions and the underlying latent processes which govern them. Our novel combination of these constituent elements represents the first model for dependent Cox point processes.

\subsection{\MakeUppercase{The Inhomogenous Poisson Process}}

For a domain $\mathcal{X} = \mathbb{R}^D$ of arbitrary dimension $D$, we may define an inhomogenous Poisson process via an intensity function $\lambda(x) : \mathcal{X} \rightarrow \mathbb{R}^+$, and a Lebesgue measure over the domain, $\ud x$. The number of events $N(\mathcal{T})$ found over a subregion $\mathcal{T} \subset \mathcal{X}$ will be Poisson distributed with parameter $\lambda_{\mathcal{T}} = \int_{\mathcal{T}} \lambda(x) \, \ud x$. Furthermore, we define $N(\mathcal{T}_i)$ to be independent random variables, where $\mathcal{T}_i$ are disjoint subsets of $\mathcal{X}$ \citep{Kingman_93_a}.

If we bound the region to be considered, and assume there are $K$ observed events, labelled as $\{x_k\}_{k=1}^K$, then the inhomogenous Poisson process likelihood function may be written as
\begin{equation}
p(\{x_k\}_{k=1}^K \mid  \lambda(x) ) = \exp \left\{- \int_{\mathcal{T}} \, \ud x \, \lambda(x) \right\} \prod\limits_{k=1}^K \lambda(x_k) \mathrm{.}
\label{eqn:intracint}
\end{equation}
\subsection{\MakeUppercase{The Sigmoidal Gaussian Cox Process}}
In order to model the intensity nonparametrically, we place a Gaussian Process \citep{Rasmussen_06_a} prior over a random scalar function $g(x) : \mathcal{X} \rightarrow \mathbb{R}$. This means that the prior over any finite set of function values $\{g(x_n)\}_{n=1}^N$ is a multivariate Gaussian distribution, defined by a positive definite covariance function $C(.,.) : \mathcal{X} \times \mathcal{X} \rightarrow \mathbb{R}$ and a mean function $m(.) : \mathcal{X} \rightarrow \mathbb{R}$. The mean and covariance function are parameterised by a set of hyperparameters, which we denote $\gamma$.

In the SGCP, a Gaussian Process is transformed into a prior over the intensity function by passing it through a sigmoid function and scaling it against a maximum intensity $\lambda^*$: $\lambda(x) = \lambda^* \sigma(g(x))$, where $\sigma(.)$ is the logistic function. This forms the basis of a generative prior, whereby exact Poisson data can be generated from $\lambda(x)$ via thinning \citep{Lewis_79_a}, which involves adding $M$ events, such that the joint point process over the $M + K$ events is homogenous with fixed rate $\lambda^*$.

As we are using an infinite dimensional proxy for $\lambda(x)$, the integral in Equation \ref{eqn:intracint} is intractable. Furthermore, using Bayes' theorem with this likelihood yields a posterior with intractable integrals in both the numerator and denominator. These challenges are overcome by making use of the generative prior, and augmenting the variable set to include the number of thinned points, $M$, and their locations, $\{\xtm\}_{m=1}^M$. This then means that the value of the intensity function need only be inferred at the $M+K$ point locations, $\mathbf{g_{M+K}} = \{g(x_k)\}_{k=1}^K \cup \{g(\xtm)\}_{m=1}^M$. Noting that $\sigma(-z)=1- \sigma(z)$, the joint likelihood over the data, function values and latent variables is
\begin{align}
p(\{x_k&\}_{k=1}^K, M, \{\xtm\}_{m=1}^M, \mathbf{g_{M+K}} \mid  \lambda^*, \mathcal{T}, \theta ) =  \notag \\
&(\lambda^*)^{M+K} \exp \left\{ -\lambda^* \mu(\mathcal{T}) \right\} \notag \\
&\times \prod\limits_{k=1}^K \sigma(g(x_k)) \prod\limits_{m=1}^M \sigma(-g(\xtm)) \notag \\
&\times \, \mathcal{GP}(\mathbf{g_{M+K}} \mid  \{x_k\}_{k=1}^K, \{\xtm\}_{m=1}^M, \gamma) \mathrm{,}\label{eqn:doableint}
\end{align}
where we have defined $\mu(\mathcal{T}) = \int\limits_\mathcal{T} \, \ud x$.

Notably, this likelihood equation does not involve any intractable integrals. This means that inference is now possible in this model, albeit subject to the cost of an augmented variable set.

\subsection{\MakeUppercase{The Convolution Process}}

The convolution process framework is an elegant way of constructing dependent output processes. Instead of assuming the typical instantaneous \citep{Teh_05_a} mixing of a set of independent processes to construct correlated output processes, we generalise to allow a blurring of the latent functions achieved via convolution with a kernel, $G(x,z)$, prior to mixing. $z$ is typically defined on the same domain as $x$. If we place a Gaussian process prior over the latent function, the output function turns out to also be a Gaussian process \citep{Alvarez_11_a}. Specifically, given $D$ dependant intensity functions $g_d(x)$ and $Q$ latent processes $u_q(x)$, (where typically $Q < D$), the stochastic component of the $d${th} intensity
is
\begin{equation}
g_d(x) = \sum \limits_{q=1}^Q \int \limits_{\mathcal{T}} G_d(x,z) u_q(z) \, \ud z \mathrm{.}
\label{eqn:convint}
\end{equation}
Given full knowledge of the latent functions, the $g_d(x)$ are independent and deterministic. The $G_d(x,z)$ encode the observed process specific characteristics, and the $u_q(z)$ can be thought of as encoding the latent driving forces.

The convolution process has strong links with the Bayesian kernel method, as described in \citep{Pillai_07_a}. This allows a function $f(x)$ on $\mathcal{X}$ to arise as
\begin{equation}
f(x) = \int_{\mathcal{X}} K(x,z) U(\ud z) \mathrm{,}
\end{equation}
where $U(\ud z) \in \mathcal{M}(\mathcal{X})$ is a signed measure on $\mathcal{X}$. The integral operator $\mathcal{L}_K  :  U(\ud z) \rightarrow f(x)$ maps the space of signed measures $\mathcal{M}(\mathcal{X})$ into $\mathcal{H}_K$, a reproducing kernel Hilbert space (RKHS) defined by the kernel, $K(x,z)$. This mapping is dense in $\mathcal{H}_K$. If we place a Gaussian process prior on the random signed measure $U(\ud z)$, rather than directly over $f(x)$, then any draw of $f(x)$ will provably lie in $\mathcal{H}_K$. As $\mathcal{H}_K$ is equivalent to the span of functions expressible as kernel integrals, this approach allows us to properly construct distributions over specific parts of function space. Using prior domain knowledge to restrict inference to plausible areas of function space is valuable, particularly for point process intensities where the likelihood linking function to data is weak and non-trivial, and we wish to only consider smooth intensities while using a sampling based inference scheme.

The convolution process is also known as a latent force model \citep{Alvarez_09_a}. In this guise, it is used to infer the solution of a differential equation when there is uncertainty in the forcing function. The convolution kernel is the Green's function of a particular differential equation, and the Gaussian process prior is placed on the driving function. This representation lets us consider the latent functions as driving forces, which are viewed through the intensity function specific convolution kernel. The convolution kernel can, for example, be used to model differing speeds of information propagation from the latent factors to each of the observed processes.

It is worth noting that any general L\'evy process prior can be used over the latent functions, however in this particular case we use a pure Gaussian process primarily for reasons of tractability.

\subsection{\MakeUppercase{Sparse Latent Functions}}

To make the model tractable, we make use of the property that the intensities are independent conditioned on the latent functions. This is made clear from the perspective of a generative model with only one latent function: we first draw a sample of the object $u(z)$, before solving the integral in equation \ref{eqn:convint}, where uncertainty about $u(z)$ is propagated through the convolution. Now instead of maintaining the full, infinite dimensional object $u(z)$, let us condition on a finite dimensional draw of $u(z)$, $u(Z) = [u(z_1),\ldots ,u(z_J)]^T$ where $Z = \{z_j\}_{j=1}^{J}$. We can then sample from $p(u(z)\mid u(Z))$, as this is a conditional Gaussian distribution, and use this function to solve the convolution integral. With multiple latent functions we can approximate each $u_q(z)$ by $\mathbb{E}[u_q(z)\mid u_q(Z)]$, replacing Equation \ref{eqn:convint} with
\begin{equation}
g_d(x) \approx \sum \limits_{q=1}^Q \int \limits_{\mathcal{T}} G_d(x,z) \mathbb{E}[u_q(z)\mid u_q(Z)] \, \ud z \mathrm{.}
\label{eqn:approxint}
\end{equation}
This is reasonable as long as each $u_q(z)$ is smooth, in the sense that it is well approximated given the covariance function and the finite dimensional sample $u_q(Z)$. In Section \ref{sec:inf}, we use the approximation in Equation \ref{eqn:approxint} along with the conditional independence assumption to build a tractable inference scheme.

\subsection{\MakeUppercase{Constructing the Model}}

Let the $Q$ latent functions $u_q(z)$ be modelled as Gaussian processes with Gaussian covariance functions such that
\begin{equation}
u_q \mid  \phi_q \sim \mathcal{GP}(0,K_q(z,z')) \mathrm{,}
\end{equation}
where $K_q(z,z')$ is simply the Gaussian kernel
\begin{equation}
K_q(z,z') = \mathcal{N}(z;z',\phi_q) \mathrm{.}
\end{equation}
We use a scaled Gaussian convolution kernel
\begin{equation}
G_{d}(x,z) = \kappa_d \dotfill \, \mathcal{N}(x;z,\theta_d) \mathrm{.}
\end{equation}
This restricts $g_d(x)$ to be at least as smooth as the random draws from $u_q(z)$. The covariance linking $u_q(z)$ to $g_d(x)$ is
\begin{align}
K_{g_d,u_q}(x,z) &= \int \limits_{\mathcal{X}} G_{d}(x,z) K_q(z,z') \ud z \notag \\
&= \kappa_d \, \mathcal{N}(x;z,\theta_d + \phi_q) \mathrm{,}
\end{align}
and the overall covariance between output functions is
\begin{align}
K_{g_d,g_{d'}}(x,x') &= \sum \limits_{q = 1}^{Q} \int \limits_{\mathcal{X}} G_{d}(x,z) \notag \\&\int \limits_{\mathcal{X}} G_{d'}(x',z') K_q(z,z') \, \ud z' \, \ud z \notag \\
K_{g_d,g_{d'}}(x,x') &= \sum \limits_{q = 1}^{Q} \kappa_d \, \kappa_{d'} \, \mathcal{N}(x;z,\theta_d + \theta_d' + \phi_q) \mathrm{.}
\end{align}

We could use this joint covariance function to construct one large joint Gaussian process over all the intensity functions. In doing this, however, the $u_q(z)$ have been implicitly integrated out, and the resulting inference problem will scale computationally as $\mathcal{O}(D^3N^3)$ with storage requirements of $\mathcal{O}(D^2N^2)$, where $N = M+K$ is the joint number of events. This is intractable for any real problem, where we would hope to leverage many dependent point processes to learn a few latent factors with minimal uncertainty.

Let us now define some additional notation: $\mathbf{K}$ denotes a covariance matrix obtained by evaluating the appropriate covariance function at all eligible pairs of data points. Subscripts determine which covariance is used and hence which inputs are valid, e.g. $\mathbf{K}_{g_d,u_q}$ denotes the cross covariance between the $d${th} output and $q${th} input function. $\mathbf{K}_{g_d,u}$ means stack the $Q$ $\mathbf{K}_{g_d,d_q}$ matrices vertically, $\mathbf{K}_{u,u}$ is a block diagonal matrix where each block corresponds to $\mathbf{K}_{u_q,u_q}$, and $\mathbf{u}$ is the result of stacking the draws from the finite dimensional Gaussians $p(u_q(Z))$ vertically.

We also define: $\phi = \{\phi_q\}_{q=1}^Q$, $\kappa = \{\kappa_d\}_{d=1}^D$, $\theta = \{\theta_d\}_{d=1}^D$, $X_d = \{x_{d,k}\}_{k=1}^{K_d} \cup \{\xt_{d,m}\}_{m=1}^{M_d}$. If we wish to allow the latent functions to be sampled at different points, then we define a separate $Z_q$ for each $u_q(z)$: $Z_q = \{z_{q,j}\}_{j=1}^{J}$. The set of inputs over all latent functions is then $Z = \{Z_q\}_{q=1}^Q$, and similarly for the intensities: $X = \{X_d\}_{d=1}^D$.

Notation in place, we determine that given the approximation in Equation \ref{eqn:approxint}, the conditional likelihood for $g_d(x)$ is
\begin{align}
p(&g_d \mid  u, Z, X_d, \kappa_d, \theta_d, \phi) = \nonumber \\
&\mathcal{N}(\mathbf{K}_{g_d,u} \mathbf{K}^{-1}_{u,u} \mathbf{u}, \, \mathbf{K}_{g_d,g_d} - \mathbf{K}_{g_d,u}\mathbf{K}^{-1}_{u,u}\mathbf{K}^T_{g_d,u}) \mathrm{.}
\label{eqn:funccov}
\end{align}
Still conditioning on the latent functions, the joint likelihood over all $D$ intensity functions is then simply
\begin{align}
p(&g_1,\ldots ,g_D \mid  u, Z, X, \kappa, \theta, \phi) = \notag \\
&\prod \limits_{d=1}^D p(g_d \mid  u, Z, X_d, \kappa_d, \theta_d, \phi) \mathrm{.}
\end{align}
Bayes' rule for Gaussians gives us the posterior over the $u_q(Z)$ as
\begin{multline}
p(u_1,\ldots ,u_Q \mid  g_1,\ldots ,g_D, Z, X, \kappa, \phi, \theta ) \\
= \mathcal{N}(u_1,\ldots ,u_Q; \mathbf{\mu}_p,\, \mathbf{\Sigma}_p) \mathrm{.}
\label{eqn:latentfunclik}
\end{multline}
Where the mean and covariance are
\begin{align}
\mathbf{\Sigma}_p &= \left[ \mathbf{K}^{-1}_{u,u} + (\mathbf{K}_{g,u}\mathbf{K}^{-1}_{u,u})^T\mathbf{D}^{-1}(\mathbf{K}_{g,u}\mathbf{K}^{-1}_{u,u}) \right]^{-1} \nonumber \\
\mathbf{\mu}_p &= \mathbf{\Sigma}_p (\mathbf{K}_{g,u}\mathbf{K}^{-1}_{u,u})^T \mathbf{D}^{-1} \mathbf{g}
\end{align}
and where $\mathbf{D} = \mathbf{K}_{g,g} - \mathbf{K}_{g,u}\mathbf{K}^{-1}_{u,u}\mathbf{K}^T_{g,u}$.

The exact form of $\mathbf{D}$ depends on the degree to which we are willing make independence assumptions in order to approximate the Gaussian processes used to model the functions. Naturally the higher the degree of approximation, the more scalable the resulting inference scheme.

Under full dependence, a single event is linked to both inter and intra-process data. We will be assuming that the dependency structure across the $g_d(x)$ is entirely contained by the latent processes, $u_q(z)$. Intuitively, this means that we maintain the full Gaussian process structure for each individual intensity function, while summarising the latent functions via a set of inducing inputs $Z = \{z_j\}_{j=1}^J$. This approximation scheme results in a functional form which is  similar to what \citet{Quinonero_05_a} call the Partial Independence (PITC) scheme. Importantly it allows inference to scale computationally as $\mathcal{O}(DN^3)$, with storage requirements of $\mathcal{O}(DN^2)$ even in the worst case scenario of $J = N$. Further approximations may be made, and these are especially useful if the number of events per process is large, however for our purposes they are not necessary. For more information on approximation methods for Gaussian processes see \cite{Quinonero_05_a} and \cite{Snelson_05_a}.

Under the PITC low rank covariance, the resulting form for $\mathbf{D}$ is: $[ \mathbf{K}_{g,g} - \mathbf{K}_{g,u}\mathbf{K}^{-1}_{u,u}\mathbf{K}^T_{g,u} ] \circ \mathbf{M}$, where $\mathbf{M} = \mathbf{I}_N \otimes \mathbf{1}_N$ and $\mathbf{1}_N$ is a $N \times N$ matrix of ones. This may be more familiar as \texttt{blkdiag}$[\mathbf{D}]$.

\section{\MakeUppercase{Inference}}\label{sec.inference}
\label{sec:inf}

For each of the $D$ point processes we need to learn $|X_d| $, $X_d$, $\kappa_d$, $\theta_d$, $\lambda^*_d$ and $g_d(x)$. For each of the $Q$ latent functions $u_q(Z)$ and $\phi_q$ must be inferred. $Z_q$ are fixed to an evenly spaced grid which is identical across the latent processes. To find posteriors over all these variables, we choose a Markov Chain Monte Carlo (MCMC) algorithm, as detailed below.

Using the PITC approximation scheme, the likelihood over the point processes factorises conditioned on the latent functions. This means that given $D$ compute units the updates associated with each point process may be made in parallel. This is important as the inference algorithm is computationally bottlenecked by operations associated with learning the locations of the thinning points, $X$.

We now give a recap of the inference scheme from the SGCP for a single point process, while listing our minor modifications. Updates for the latent functions are then given, conditioning on the $D$ intensity functions.

\subsection{\MakeUppercase{Learning the Intensity Function}}

Recalling Equation \ref{eqn:doableint}, three kinds of Markov transitions are used to draw from this joint distribution: 1) Sampling the number of thinned points, $M$. 2) Sampling the locations of the thinned events, $\{\xtm\}_{m=1}^M$. 3) Resampling the intensity function, $\mathbf{g_{M+K}}$.

Metropolis-Hastings is used to sample $M$. The probability of insertion/deletion is parameterised by a Bernoulli proposal function: $b(K,M) : \mathbb{N} \times \mathbb{N}\rightarrow (0,1)$, where the parameter has been arbitrarily set to $\frac{1}{2}$. If an insertion is required, a new $x_{M+1}$ is drawn uniformly and at random from $\mu(\mathcal{T})$, and $g(x_{M+1})$ is drawn from the Gaussian process conditioned on the current state. A deletion results in a thinned event $\xtm$ being removed at random from $\{\xtm\}_{m=1}^M$. The overall transition kernels $q$, and Metropolis-Hastings acceptance ratios, $a$, are:
\begin{eqnarray}
&q_{ins}(M+1 \leftarrow M) = \label{eqn:samplethinbeg} \\
&\frac{b(K,M)}{\mu(\mathcal{T})} \mathcal{GP}(g(\xt_{M+1}) \mid  \{\xtm\}_{m=1}^M, \mathbf{g_{M+K}}) \mathrm{,} \nonumber
\end{eqnarray}
\begin{equation}
a_{ins} = \frac{(1 - b(K,M+1))\mu(\mathcal{T})\lambda^*}{(M+1)b(K,M)(1 + \exp(g(\xt_{M+1})))} \mathrm{,}
\end{equation}
\begin{equation}
q_{del}(M-1 \leftarrow M) = \frac{1 - b(K,M)}{M}
\end{equation}
\begin{equation}
a_{del} = \frac{M b(K,M-1)(1+\exp(g(\xtm)))}{(1-b(K,M))\mu(\mathcal{T}) \lambda^*} \mathrm{.}
\label{eqn:samplethinend}
\end{equation}
Sampling the locations of the thinned events also makes use of the Metropolis criterion. For each event $\xtm$ a move to $\hat{x}_m$ is proposed via a Gaussian proposal density. A function value $g(\hat{x}_m)$ is then drawn conditioned on the state with $g(\xtm)$ removed, denoted $\mathbf{g_{M_-+K}}$. This gives the move acceptance ratio
\begin{equation}
a_{move} = \frac{q_{move}(\xtm \leftarrow \hat{x}_m)(1 + \exp(g(\xtm)))}{q_{move}(\hat{x}_m \leftarrow \xtm)(1 + \exp(g(\hat{x}_m)))} \mathrm{.}
\label{eqn:samplejitterbeg}
\end{equation}
where $q_{move}$ is the proposal distribution. We use a symmetric Gaussian proposal
\begin{equation}
q_{move}(\hat{x}_m \leftarrow \xtm) = \mathcal{N}\left(0,\frac{\mu(\mathcal{T})}{100}\right) \mathrm{.}
\label{eqn:samplejitterend}
\end{equation}
To sample the function we opt to use Elliptical Slice Sampling \citep{Murray_10_a}. This is an algorithm specifically designed for sampling from high dimensional, highly correlated, Gaussian process posteriors. The log conditional posterior over function values is
\begin{eqnarray}
&\ln \, p(\mathbf{g_{M+K}} \mid  M, \{x_k\}_{k=1}^K, \{\xtm\}_{m=1}^M, \gamma ) = \nonumber \\
&-\frac{1}{2} \mathbf{g_{M+K}} \mathbf{\Sigma^{-1}} \mathbf{g_{M+K}} - \sum \limits_{k=1}^K \ln(1 + \exp(-g(x_k))) \nonumber \\
&-\sum\limits_{m=1}^{M} \ln(1 + \exp(g(\xtm))) + const.
\label{eqn:samplefunc}
\end{eqnarray}
In our case, $\mathbf{\Sigma}$ is equal to the covariance in Equation \ref{eqn:funccov}, and naturally for each iteration we perform all the above updates in parallel for each observed point process, conditioned on the latent functions.

To infer the posteriors over the Gaussian process hyperparameters, we use Hamiltonian Monte Carlo (HMC) \citep{Duane_87_a,Neal_10_a}, with log-normal priors over each hyperparameter. By placing a Gamma prior with shape $\alpha$ and inverse scale $\beta$ over $\lambda^*$, we infer the posterior conditioned on the thinned and true points using a Gibbs update as follows:
\begin{eqnarray}
\alpha_{post} = \alpha + K + M, & \beta_{post} = \beta + \mu{\mathcal{T}} .
\label{eqn:samplelambda}
\end{eqnarray}

\subsection{\MakeUppercase{Learning the Latent Functions}}

Conditioning on the point process intensity functions, $g_d(x)$, the latent functions are dependent, with conditional posterior distribution given by Equation \ref{eqn:latentfunclik}.

Having drawn new values for each of the $u_q(Z_q)$, we can update the $\phi_q$ using a metropolis-hastings step under the following log conditional posterior which is
\begin{eqnarray}
&\ln p(\phi_q \mid  u_q(Z_q), Z_q) = -\frac{1}{2}u_q(Z_q)\mathbf{K}^{-1}_{u_q,u_q}\frac{1}{2}u_q(Z_q) \nonumber \\
&  - \frac{1}{2}\log\det(\mathbf{K}_{u_q,u_q}) + \text{const.}
\label{eqn:latenthyp}
\end{eqnarray}

The overall procedure is summarised in algorithm \ref{alg:inference}.

\algblock{ParFor}{EndParFor}
\begin{algorithm}[tb]
   \caption{MCMC Scheme}
   \label{alg:inference}
\begin{algorithmic}
   \State {\bfseries Input:} $\{X_k\}_{k=1}^K$, priors.
   \Repeat
   \ParFor{ $d=1$ {\bfseries to} $D$}
   \State Sample thinned events: Equations \ref{eqn:samplethinbeg} $\rightarrow$ \ref{eqn:samplethinend}
   \State Sample locations: Equations \ref{eqn:samplejitterbeg} $\rightarrow$ \ref{eqn:samplejitterend}
   \State Sample function: Equation \ref{eqn:samplefunc}
   \State Sample hyperparameters: Equation \ref{eqn:samplefunc}
   \State Sample $\lambda^*$: Equation \ref{eqn:samplelambda}
   \EndParFor
   \State Sample latent functions: Equation \ref{eqn:latentfunclik}
  \State Sample latent hyperparameter: Equation \ref{eqn:latenthyp}
  \Until{\emph{convergence} is \emph{true}}
\end{algorithmic}
\end{algorithm}

\section{\MakeUppercase{Adaptive Thinning}}\label{sec.adaptive}
In higher dimensional spaces, data is typically concentrated into small, high density sub-domains. Under the current methodology, we must thin the entire empty space to a uniform concentration which matches that of the most dense subregion. If we wish to use Gaussian Process intensities this rapidly becomes infeasible, even under the most radical of sparse approximations \citep{Snelson_05_a}.

Our novel solution to this problem is to model the upper bounded intensity over the space using a piece-wise constant function, where each section takes a fractional proportion of the global upper bound, $\lambda^*$. This preserves the tractability of the integrals in the likelihood and posterior, and does not violate any of the properties of the point process, while simultaneously allowing empty regions to be thinned to a far lower average density.

\begin{figure}
\centering
\includegraphics[width=\columnwidth, trim = 18mm 15mm 10mm 5mm, clip]{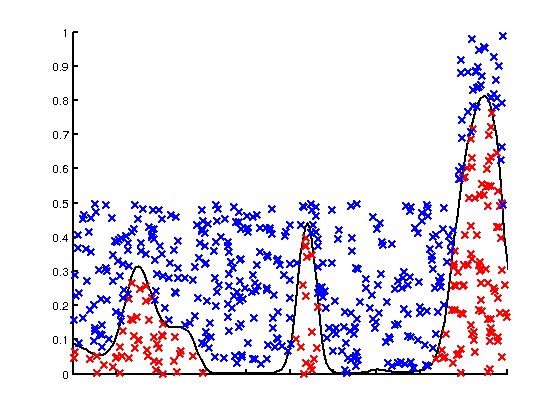}
\caption{Graphical representation of adaptive thinning: Blue crosses indicate thinned points, red crosses represent data. The black line shows the intensity function. Each point is accepted as data with probability given by $\sigma(g(x_n))$. Fewer thinned points are required in areas of half maximum bound.}
\label{fig:athin}
\end{figure}

Consider Figure \ref{fig:athin}: this shows both the data and the thinned points, where for the left three quarters of the plot the maximum rate does not exceed 50\% of $\lambda^*$. Let us assume we allow the maximum rate to take one of two values for each datapoint: $\frac{1}{2} \lambda^*$ and $\lambda^*$. For each new thinned point we sample an intensity function value, before also sampling an upper bound for the rate from the available levels. This upper bound is at least as great as the current function evaluation at that point.

In this manner we hope to infer that for the majority of Figure \ref{fig:athin}, the rate may be happily upper-bounded by half the global maximum rate, $\lambda^*$, and hence the bulk of the space may be thinned to a significantly lower density. As a result, the computational burden incurred will be significantly reduced, as far fewer expensive points need be incorporated into our GP.

In our particular implementation, we fix \emph{a-priori} a set of $B$ possible maximum rate `levels':
\begin{equation}
L = \{l_i \in  (0, 1] | l_i < l_{i+1},l_{B} = 1\}_{i=1}^B.
\end{equation}
We then augment the variable set to include for each thinned point $\xtm$ which rate level $r_m \in \{1\hdots B\}$ it is currently assigned, where we set $r_m$ such that $\sigma(g(\xtm)) \leq l_{r_m}$. This causes the probability of seeing a thinned point $\xtm$ under the sigmoid GP to become
\begin{equation}
	p( \xtm | r_m ) = \frac{l_{r_m} - \sigma(g(\xtm))}{l_{r_m}} \; \mathrm{,}
\end{equation}
while the probability of a non-thinned point remains unchanged. Using this relationship we modify the Metropolis acceptance criteria which now become
\begin{equation}
a_{ins} = \frac{(1 - b(K,M+1))\mu(\mathcal{T})\lambda^*l_{r_{M+1}}p(\xt_{M+1}|r_{M+1})}{(M+1)b(K,M)} \mathrm{,}
\end{equation}
\begin{equation}
a_{del} = \frac{M b(K,M-1))}{(1-b(K,M))\mu(\mathcal{T}) \lambda^*l_{r_m}p(\xtm|r_{m})} \mathrm{,}
\end{equation}
as well as the likelihood function for $p(g(\mathbf{X_{M+K}}))$, Equation \ref{eqn:samplefunc}.

In principle this scheme could slow mixing, since the function is constrained to lie below the maximum level at each point. By ensuring that there is always some slack, $s$, between the function and the rate level assigned we find that mixing is hardly affected. The slack is incorporated by assigning the rate as follows:
\begin{equation}
r_m \leftarrow \left\{ \begin{array}{ll}
\underset{r} {\mathrm{argmax}}\left\{ \sigma(g(\xtm)) \leq l_r \times s \right\}, & \sigma(g(\xtm)) \leq s \\
1, & \mathrm{otherwise.}
\end{array} \right. \\
\end{equation}

We used $s$ = 0.9. The rate levels can also change at each iteration during the `move' step when we compute the new rate level for jittered points and we compose the acceptance criteria as the product of the insertion and deletion criteria $a_{move} = a_{ins} \times a_{del}$.

Finally, when re-sampling $\lambda^*$ we must compute an estimate of the total number of points under a single rate uniformisation of the space. This estimate is readily available because the number of points (including observed data points) with rate $r$ is a Monte-Carlo integral of the proportion of space thinned to rate level $l_r$. Since the number of points in each region is scaled by $l_r$, the estimated total is
\begin{equation}
\hat{N}_{tot} = \sum_{k=1}^K \frac{1}{l_{\bar{r}_k}} + \sum_{m=1}^M \frac{1}{l_{r_m}} \mathrm{,}
\end{equation}

where $\bar{r}_k$ is the notional rate of observed data computed exactly as for the thinned points. The posterior value $\alpha_{post}$ is therefore $\alpha + \hat{N}_{tot}$.

To validate adaptive thinning, we return to the original SGCP and modify it in the manner described above. We perform two experiments: The first in 1D and the second in 2D. In both cases we use 10 known random intensity functions to generate event data: In the 1D case we sample 15 random datasets per function, while in the 2D case we generate 10. One dataset is used to learn the model, the rest are held out for testing purposes. Two metrics of performance are used: L2-norm error as measured against the true intensity function, and average predictive log-likelihood across all held out test datasets.

In 1D, we run each model for 6 minutes total compute time, in 2D we allow 20 minutes total. In both cases half the time is allocated to burn-in. In 1D the single rate method achieved roughly one sample per second, with the two rate case yielding just under four samples per second, and the four rate approach giving just under 8 samples per second. In 2D the number of points required was larger in all cases, with the multi-rate approach buying a factor of two speedup. 

The results, (given in Tables \ref{tab:1drms} through \ref{tab:2dll}), show that in almost all cases the multi-level approach performs best across both metrics. It is observed that typically the original, homogenous rate approach performs worst of all.

\begin{table*}[htp]
\begin{minipage}{0.45\textwidth}
\caption{\centering{One dimensional adaptive thinning L2-norm function error. Bold is best.}}
\label{tab:1drms}
\begin{center}
\begin{tabular}{llll}
\toprule
Function & \multicolumn{3}{c}{L2 Norm Error} \\
  (1D)   & Original & 2 Rates & 4 Rates  \\
\midrule
1 & 11.6 & 13.0 & \bf{7.1} \\
2 & 14.6 & 10.5 & \bf{7.2} \\
3 & 10.6 & 5.0 & \bf{4.9} \\
4 & 10.5 & \bf{4.7} & 5.1 \\
5 & 12.4 & \bf{10.1} & 10.4 \\
6 & 11.1 & 8.2 & \bf{7.6} \\
7 & \bf{12.0} & 13.7 & 12.5 \\
8 & 13.0 & \bf{12.0} & 12.8 \\
9 & 19.6 & \bf{16.4} & 28.8 \\
10 & 31.4 & \bf{27.2} & 32.6 \\
\bottomrule
\end{tabular}
\end{center}
\end{minipage}
\hspace{1cm}
\begin{minipage}{0.45\textwidth}
\caption{\centering{One dimensional adaptive thinning average predictive log-likelihood on 14 held out datasets. Bold is best.} }
\label{tab:1dll}
\begin{center}

\begin{tabular}{llll}
\toprule
Function & \multicolumn{3}{c}{Predictive Log-Likelihood} \\
  (1D)   & Original & 2 Rates & 4 Rates  \\
\midrule
1 & 373.8 & 381.8 & \bf{388.2} \\
2 & 626.1 & 644.2 & \bf{650.7} \\
3 & 274.2 & 285.1 & \bf{288.0} \\
4 & 435.5 & \bf{457.7} & 456.0 \\
5 & 877.0 & 885.8 & \bf{889.5} \\
6 & 995.4 & 1006.6 & \bf{1013.4} \\
7 & 753.0 & \bf{763.3} & 760.6 \\
8 & 522.3 & \bf{531.2} & 528.3 \\
9 & 1840.6 & \bf{1852.9} & 1826.0 \\
10 & 2328.1 & \bf{2365.8} & 2349.8 \\
\bottomrule
\end{tabular}
\end{center}
\end{minipage}
\end{table*}

\begin{table*}[htp]
\begin{minipage}{0.45\textwidth}
\caption{\centering{Two dimensional adaptive thinning L2-norm function error. Bold is best.}}
\label{tab:2drms}
\begin{center}
\begin{tabular}{llll}
\toprule
Function & \multicolumn{3}{c}{L2 Norm Error} \\
  (2D)   & Original & 2 Rates & 4 Rates  \\
\midrule
1 & 13.3 & 13.4 & \bf{11.3} \\
2 & 14.3 & \bf{14.3} & 14.7 \\
3 & 13.5 & 14.7 & \bf{13.5} \\
4 & 12.5 & 12.9 & \bf{12.0} \\
5 & 17.9 & \bf{16.6} & 17.8 \\
6 & \bf{14.4} & 16.2 & 15.3 \\
7 & 15.1 & \bf{13.7} & 17.5 \\
8 & 14.6 & \bf{14.4} & 14.8 \\
9 & 18.3 & 17.1 & \bf{15.6} \\
10 & 15.4 & \bf{12.9} & 13.6 \\
\bottomrule
\end{tabular}
\end{center}
\end{minipage}
\hspace{1cm}
\begin{minipage}{0.45\textwidth}
\caption{\centering{Two dimensional adaptive thinning average predictive log-likelihood on 9 held out datasets. Bold is best.}}
\label{tab:2dll}
\begin{center}
\begin{tabular}{llll}
\toprule
Function & \multicolumn{3}{c}{Predictive Log-Likelihood} \\
  (2D)   & Original & 2 Rates & 4 Rates  \\
\midrule
1 & 2039.6 & 2080.6 & \bf{2203.0} \\
2 & 2757.9 & 2758.1 & \bf{2829.3} \\
3 & 2753.2 & 2689.5 & \bf{2827.2} \\
4 & 2803.2 & 2784.0 & \bf{2933.6} \\
5 & 2532.4 & \bf{2663.0} & 2572.1 \\
6 & \bf{3098.2} & 3040.5 & 3054.4 \\
7 & 3157.5 & \bf{3259.8} & 3075.2 \\
8 & 2086.9 & \bf{2101.0} & 2087.4 \\
9 & 5018.1 & 5146.6 & \bf{5185.3} \\
10 & 2008.1 & \bf{2205.0} & 2174.0 \\
\bottomrule
\end{tabular}
\end{center}
\end{minipage}
\end{table*}

\section{\MakeUppercase{Empirical Results}}
As this is the first model for structured point process data, the approach is initially validated on a synthetic dataset. It is then compared to both the independent SGCP, as well as a state of the art Kernel Density Estimator \citep{Botev_10_a} on two real datasets.

\subsection{\MakeUppercase{Synthetic Data}}
\begin{figure*}[htp]
\centering
\begin{minipage}{0.48\textwidth}
\centering
\includegraphics[width=\columnwidth, trim = 44mm 26mm 31mm 17mm, clip]{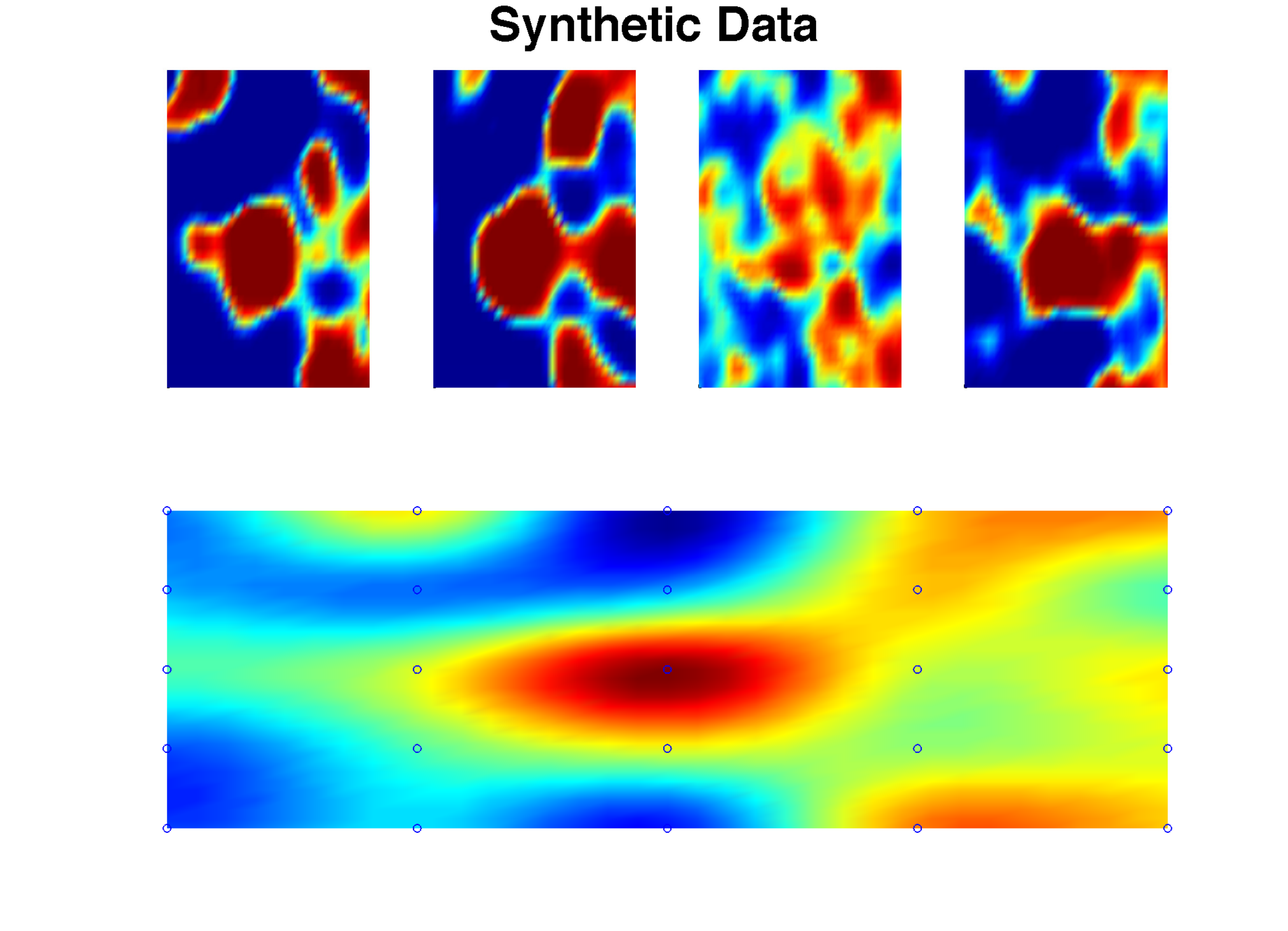}
\caption{\centering{Synthetic functions (not showing the sampled events).}}\label{fig:gendata}
\end{minipage}
\hfill
\begin{minipage}{0.48\textwidth}
\centering
\includegraphics[width=\columnwidth, trim = 38mm 22mm 26mm 15mm, clip]{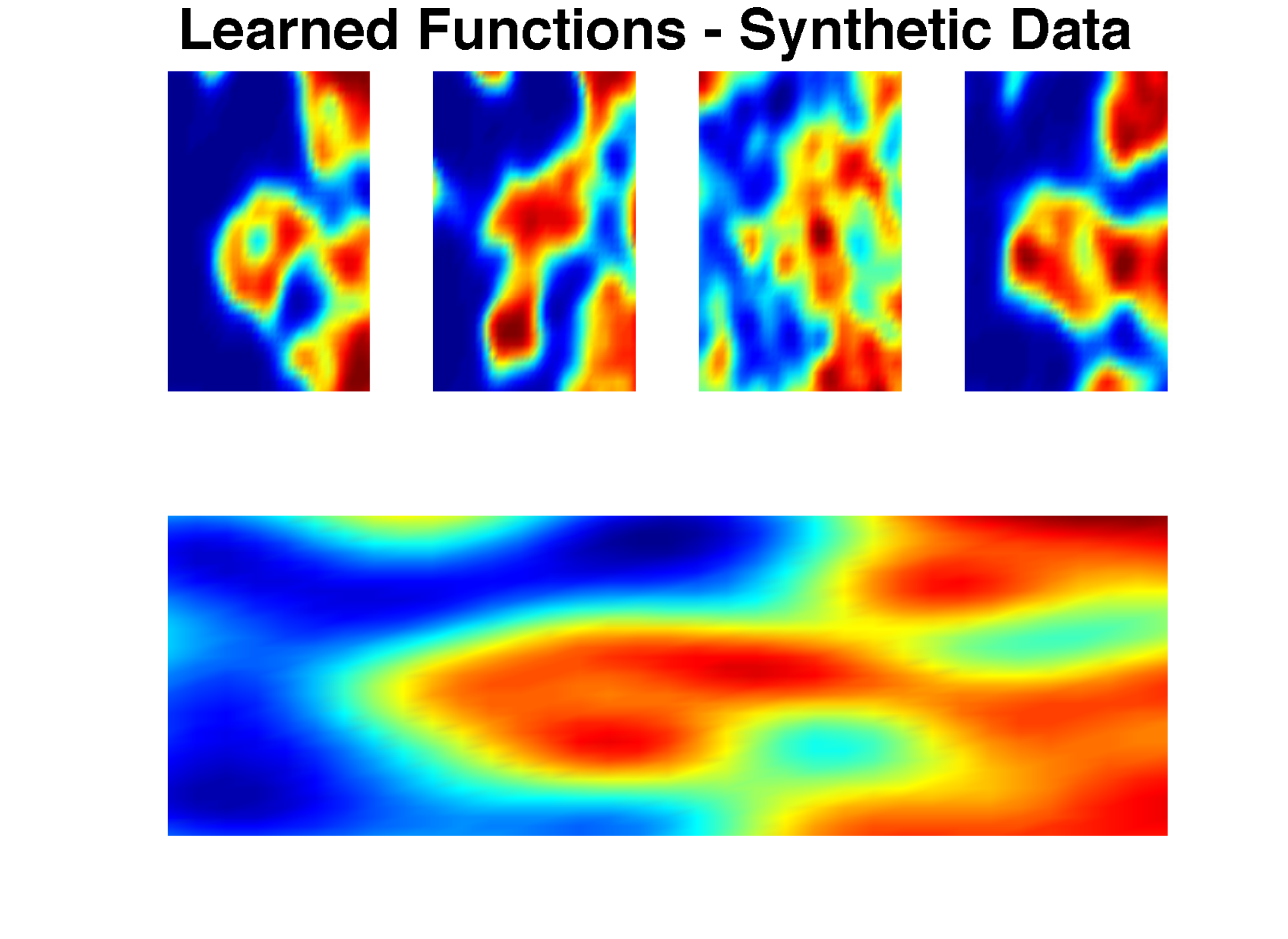}
\caption{\centering{Learned functions using 3 rate levels: $\frac{1}{4}\lambda^*$, $\frac{1}{2}\lambda^*$, $\lambda^*$.}}\label{fig:genresults}
\end{minipage}
\end{figure*}

Using the convolution process, we sample four intensity functions, using those to sample event data. The variety of intensities which may be observed given a single latent function is notable in Figure \ref{fig:gendata}.

We then average over 2000 iterations after it was determined convergence had been achieved. The resulting learned intensity functions are shown in Figure \ref{fig:genresults}.

It is reassuring that the original latent function is well recovered given only four observed event processes.

\subsection{\MakeUppercase{Real Data}}\label{sec.realdata}

Two datasets were selected to test the model, both of which we considered were likely to exhibit a dependency structure which could be well captured by the convolution
process.

\begin{itemize}
\item British politicians (MPs) tweet times during the week of Nelson Mandela's death (02/12/13-08/12/13). These were obtained using the Twitter API. Here we considered that there would naturally be a daily periodicity, however, it is not unreasonable to further postulate that some MPs may concentrate their twitter activity into a smaller segment of the day. This behaviour should be well captured by the convolution process.
\item NBA player shot profiles for the 2013-2014 season, scraped from the NBA website. Here we select a diverse subset of four players: Blake Griffin, Damien Lillard, DeMar DeRozen, and Arron Affalo. It was supposed that it might be possible for a single latent function to be blurred to represent a variety of player positions and styles.
\end{itemize}

\subsubsection{Twitter Data Results}
Four MPs active on twitter were selected at random. Here on we call them MPs A, B, C, and D. We select data from the period covering 02/12/13 through 13/12/13, and randomly partition each dataset into 75\% training data, with the remainder being used to evaluate predictive test log-likelihood.

\begin{figure}[htp]
\centering
\includegraphics[width=\columnwidth, trim = 20mm 0mm 10mm 0mm, clip]{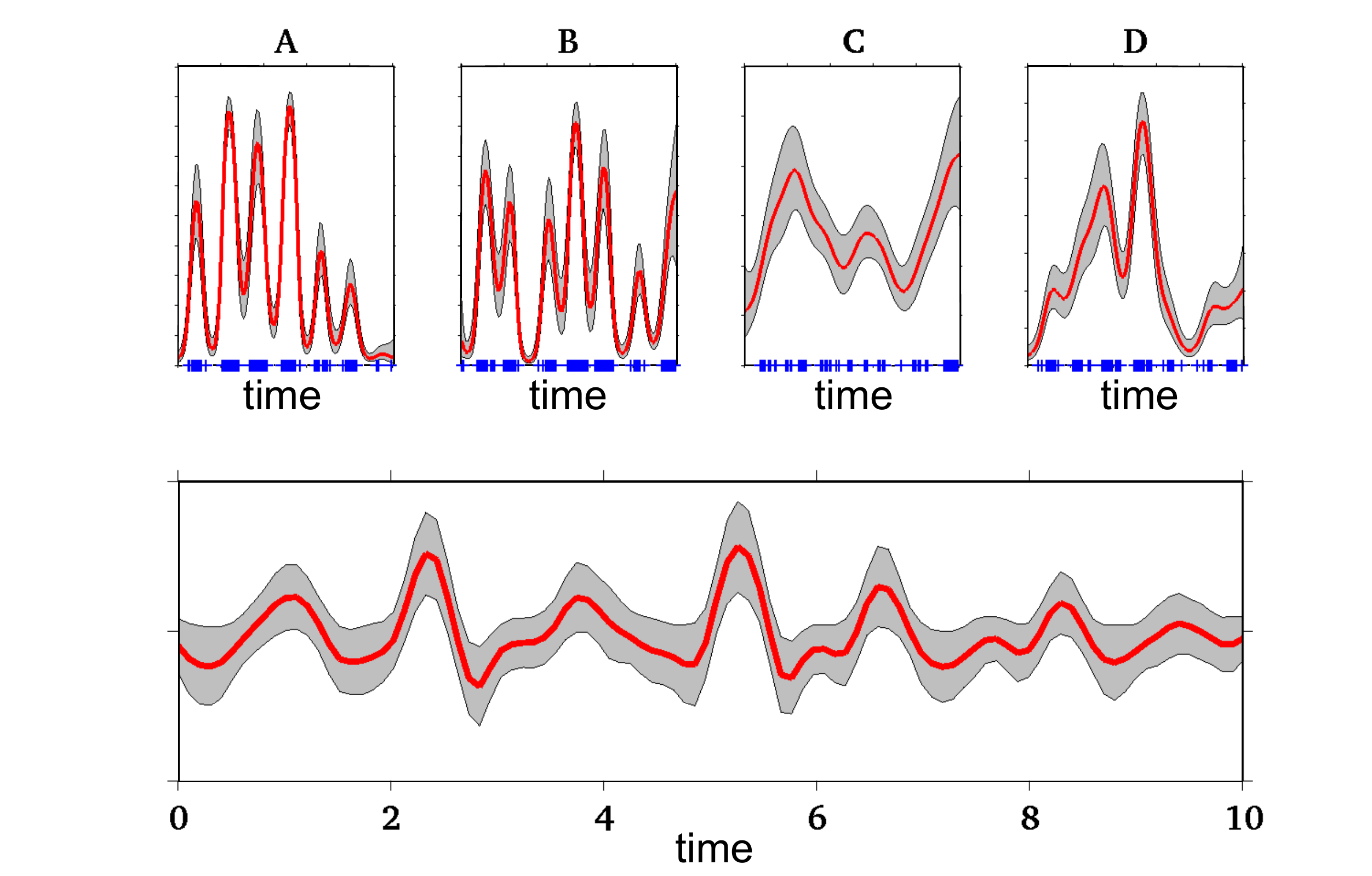}
\caption{Learned intensities over four MP's tweet data (A, B, C, D); learned latent function at bottom. Actual data shown in blue.}
\label{fig:tweetdata}
\end{figure}

Figure \ref{fig:tweetdata} depicts the average learned intensity functions for each MP (red line), along with the one standard deviation bars (grey shading) derived from the function samples. The bottom plot depicts the learned latent driving function in the same manner.

The latent function clearly shows a strong daily period, particularly evident during the working week (02/12/13 was a Monday---corresponding to `1' in Figure \ref{fig:tweetdata}). Furthermore, the largest two peaks in activity occur on the 3rd and the 5th of December. Potential contributing factors to these two spikes include a public sector strike, and the death of Nelson Mandela respectively.

\begin{figure*}[htp]
\begin{minipage}{0.45\textwidth}
\centering
\includegraphics[width=\columnwidth, trim = 23mm 18mm 20mm 5mm, clip]{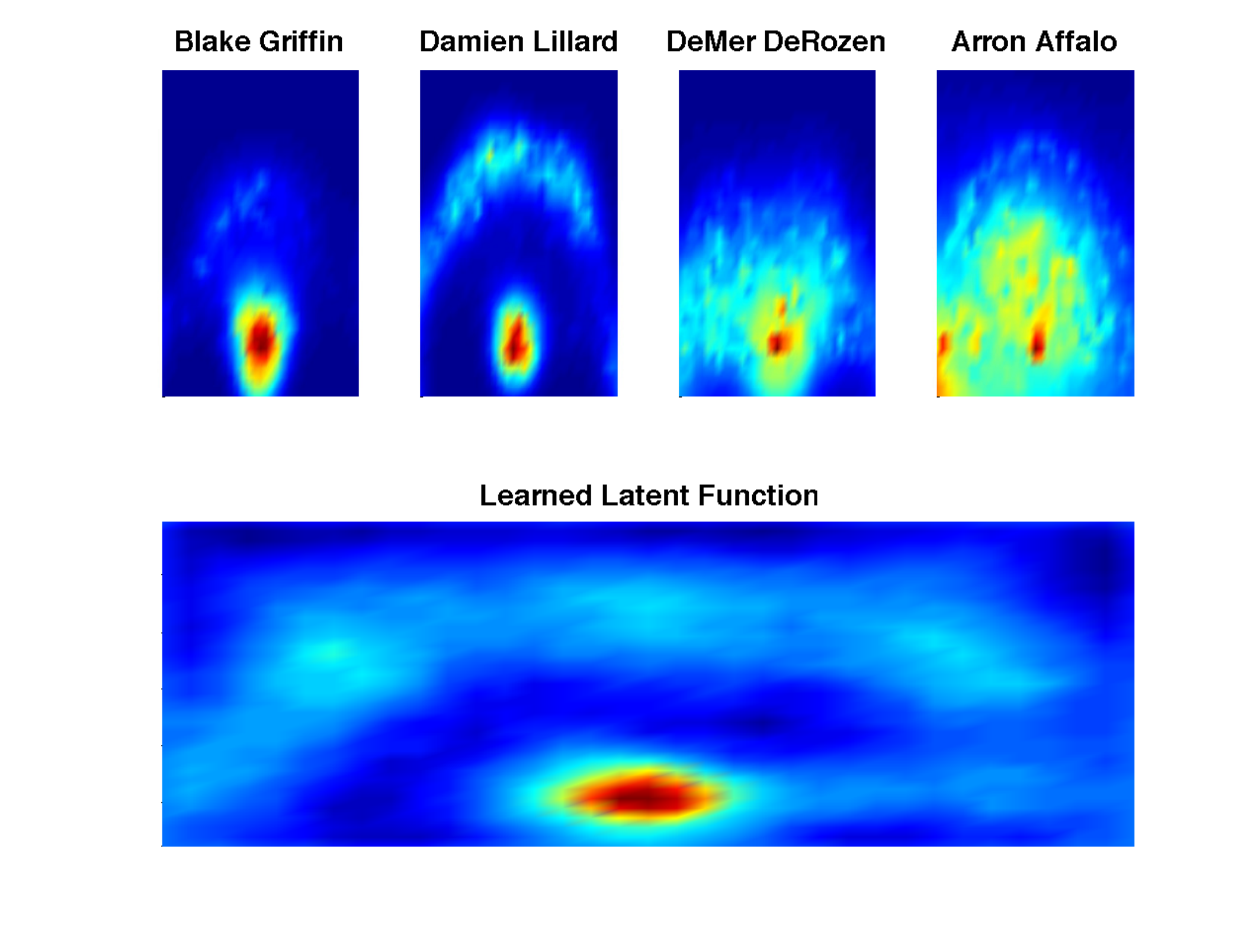}
\caption{Learned basketball intensity functions using 3 rate levels: $\frac{1}{4}\lambda^*$, $\frac{1}{2}\lambda^*$, and $\lambda^*$. }
\label{fig:bbresults}
\end{minipage}
\hspace{1cm}
\begin{minipage}{0.45\textwidth}
\centering
\includegraphics[width=\columnwidth, trim = 30mm 5mm 25mm -12mm, clip]{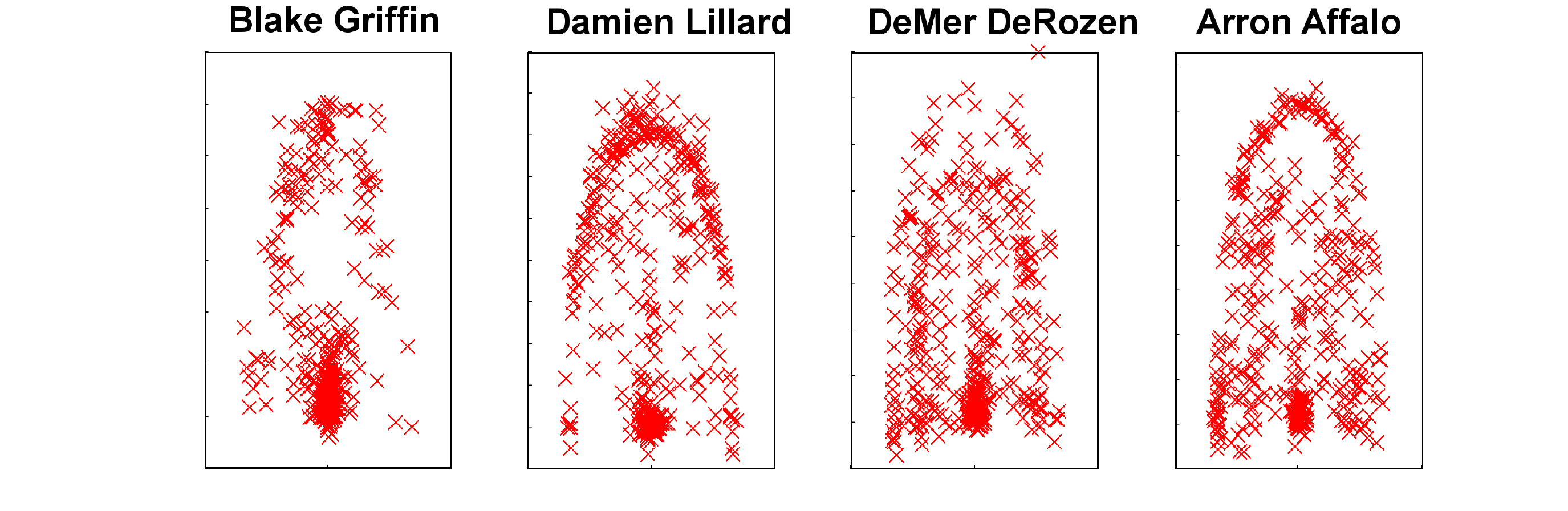}
\label{fig:bbdata}
\label{tab:bbll}
\begin{center}
\begin{tabular}{llll}
\toprule
Player   & \multicolumn{3}{c}{Predictive Log-Likelihood} \\
         & KDE & SGCP & Ours  \\
\midrule
Blake Griffin & -121.8 & 335.6 & \bf{374.7} \\
Damien Lillard & -22.6 & 231.2 & \bf{395.3} \\
DeMer DeRozen & -2.9 & 253.1 & \bf{410.7} \\
Arron Affalo & -260.7 & -76.7 & \bf{84.2} \\
\bottomrule
\end{tabular}
\caption{Basketball ball shot data (top) and predictive log-likelihood for held out basketball data across models (bottom).}\label{fig.bbresults}
\end{center}
\end{minipage}
\end{figure*}

Table \ref{tab:twll} gives predictive log-likelihood for the held out data, again evaluated across three approaches: An intensity function learned via Kernel Density Estimation (KDE) \citep{Botev_10_a}, the SGCP \citep{Adams_09_a}, and our own structured approach. Both the SGCP and our own approach use two maximum rate levels for adaptive thinning. Each intensity function is modelled using an independent SGCP/KDE. The structured approach performs vastly better, suggesting that it is highly appropriate for this type of data.

\begin{table}[h]
\caption{Predictive log-likelihood for held out twitter data across models.}
\label{tab:twll}
\begin{center}
\begin{tabular}{llll}
\toprule
MP   & \multicolumn{3}{c}{Predictive Log-Likelihood} \\
         & KDE & SGCP & Ours  \\
\midrule
A & 177.1 & 176.6 & \bf{469.5} \\
B & -1.3 & 89.6 & \bf{412.9} \\
C & -5.4 & 49.6 & \bf{283.4} \\
D & -67.7 & 38.7 & \bf{293.7} \\
\bottomrule
\end{tabular}
\end{center}
\end{table}

\subsubsection{Basketball Data}

For the basketball point shot data, the approach performed particularly well. Each player had around 600 attempted shots, of which we used 400, holding out the rest as test data. We used 3 rate boundaries: $\frac{1}{4}\lambda^*$, $\frac{1}{2}\lambda^*$, $\lambda^*$, and once again averaged over 2000 samples after convergence. We compare predictive log-likelihood to both the SGCP model (using the same set of rate boundaries) and using a rate function estimated via a state of the art KDE by \cite{Botev_10_a}.

Figure \ref{fig:bbresults} depicts the resulting intensity functions. It is clear that the latent function represents a general view of the court hotspots, the hoop and three pointer line are clearly demarcated. Furthermore the intensity functions for each player strongly match what would be expected given their playing style---e.g. Arron Affalo is a `shooting guard' who is expected to spend the majority of his time inside the three pointer line, but has a propensity to shoot from the bottom left of the court. These effects are clearly visible on the heat map, less so on the data (see Figure \ref{fig:bbdata}).

As is clearly demonstrated in Table \ref{fig.bbresults}, our structured approach to modelling the basketball point data in a fully Bayesian fashion yields a huge improvement over both the independent SGCP as well as a modern kernel density estimator. Another point worth making is that due to the  high data density around the hoop for each player, the traditional approach of thinning (used here as well as in the SGCP) would be prohibitively computationally expensive. We are only able to test on this data due to the method of adaptive thinning introduced in this paper.

\section{\MakeUppercase{Conclusion}}
We have introduced a fully generative model for dependent point processes, alongside an efficient, parallelised inference scheme. We have shown the appropriateness of this model on two real datasets, and introduced a new adaptation of thinning which allows the model to scale to larger datasets and in particular higher dimensional spaces. Future work entails investigating the appropriateness of manually introducing known latent drivers, exploring multiple latent functions, and replacing the MCMC inference scheme with one based on stochastic variational inference \citep{Hensman_13_a}.

\subsubsection*{Acknowledgements}
Tom Gunter is supported by UK Research Councils. Chris Lloyd is funded by a DSTL PhD Studentship.

\pagebreak

\subsubsection*{References}

\begingroup
\renewcommand{\section}[2]{}%
\bibliographystyle{plainnat}
\bibliography{paper.bib}
\endgroup

\end{document}